\PassOptionsToPackage{table,xcdraw,dvipsnames}{xcolor}

\documentclass[letterpaper]{article} 

\usepackage[preprint]{aaai2027} 
\usepackage{times}              
\usepackage{helvet}             
\usepackage{courier}            
\usepackage[hyphens]{url}       
\usepackage{graphicx}           
\usepackage{natbib}             
\usepackage{caption}            

\usepackage[utf8]{inputenc}
\usepackage{xcolor}
\usepackage{microtype}
\usepackage{nicefrac}

\usepackage{amsmath}
\usepackage{amssymb}

\usepackage{booktabs}
\usepackage{multirow}
\usepackage{array}
\usepackage{makecell}
\usepackage{colortbl}
\usepackage{siunitx}
\usepackage{subcaption}

\usepackage{algorithm}
\usepackage{algpseudocode}

\usepackage[shortlabels]{enumitem}

\usepackage{pifont}
\usepackage{bbding}
\usepackage{fontawesome5}
\usepackage{xspace}

\usepackage[most]{tcolorbox}

\usepackage{minted}
\usepackage{listings}
\usepackage{newfloat}

\usepackage{cleveref}

\newcommand{\method}{%
    {\fontfamily{pcr}\selectfont\textbf{Trident}}\xspace
}

\newcommand{\methodname}[1]{%
    {\fontfamily{pcr}\selectfont #1}\xspace
}

\newcommand{\modelname}[1]{%
    {\footnotesize\fontfamily{cmss}\selectfont #1}\xspace
}

\definecolor{darkgreen}{RGB}{0,128,0}
\definecolor{softred}{RGB}{180,30,30}
\definecolor{darksalmon}{rgb}{0.91,0.59,0.48}
\definecolor{mygrey}{gray}{0.4}

\newcommand{\blue}[1]{%
    $_{\color{BlueGreen}\downarrow #1}$
}

\newcommand{\red}[1]{%
    $_{\color{RedOrange}\uparrow #1}$
}

\newcommand{\hlfirst}[1]{%
    \colorbox[HTML]{CFE2FF}{#1}
}

\newcommand{\hlsecond}[1]{%
    \colorbox[HTML]{E8F1FF}{#1}
}

\DeclareCaptionStyle{ruled}{
    labelfont=normalfont,
    labelsep=colon,
    strut=off
} 

\floatstyle{ruled}
\newfloat{listing}{tb}{lst}{}
\floatname{listing}{Listing}

\title{Read, Route, Patch: Robust Test-Time Evolution for LLM Agents}

\newcommand{\affmark}[1]{\textsuperscript{\ensuremath{#1}}}

\author{
    Kaijie Chen\equalcontrib\affmark{\eta},
    Liang Yan\equalcontrib\affmark{\alpha},
    Yilan Wei\affmark{\beta},
    Jiachen Luo\affmark{\gamma},
    Zhihao Dou\affmark{\delta},\\
    Jiashu Qu\affmark{\epsilon},
    Haoran Liu\affmark{\zeta},
    Junli Wang\affmark{\eta}
}

\affiliations{}

\begin{document}

\maketitle

\begingroup
\renewcommand{\thefootnote}{}
\footnotetext{%
    \affmark{\alpha} Fudan University;\quad
    \affmark{\beta} Northwestern University;\quad
    \affmark{\gamma} Queen Mary University of London;\quad
    \affmark{\delta} Duke University;\quad
    \affmark{\epsilon} University of California, Irvine;\quad
    \affmark{\zeta} The Chinese University of Hong Kong.
}
\addtocounter{footnote}{-1}
\endgroup

\begin{abstract}
Large Language Model (LLM)-based agents can improve long-horizon interactive decision-making through test-time evolution, yet existing methods face three information bottlenecks: memory retrieval overemphasizes textual similarity, candidate harnesses are often behaviorally redundant, and whole-harness rewriting may overwrite useful evidence or introduce cross-component inconsistency. We introduce \textbf{\method}, which improves experience utilization, behavioral exploration, and harness evolution through boundary-aware memory reading, diversity-aware candidate routing, and order-aware insertion-based updating, respectively. We evaluate \method on five benchmarks spanning text games, web interaction, and coding tasks. Across \modelname{qwen-32B}, \modelname{mistral-small-3.2-24B}, and \modelname{gpt-oss-20B}, \method achieves the best average performance and consistently outperforms non-learning, memory-based, reflection-based, and automated prompt optimization baselines. Compared with the strongest baseline, APEX, \method yields relative average gains of $30.7\%$, $30.8\%$, and $38.3\%$ on the three backbones, respectively. Ablation and sensitivity studies further show that the three designs provide complementary gains and remain robust to parameter variations.
\end{abstract}

\begin{figure*}[t]
    \centering
    \makebox[\textwidth][c]{%
        \includegraphics[height=0.29\textwidth]
        {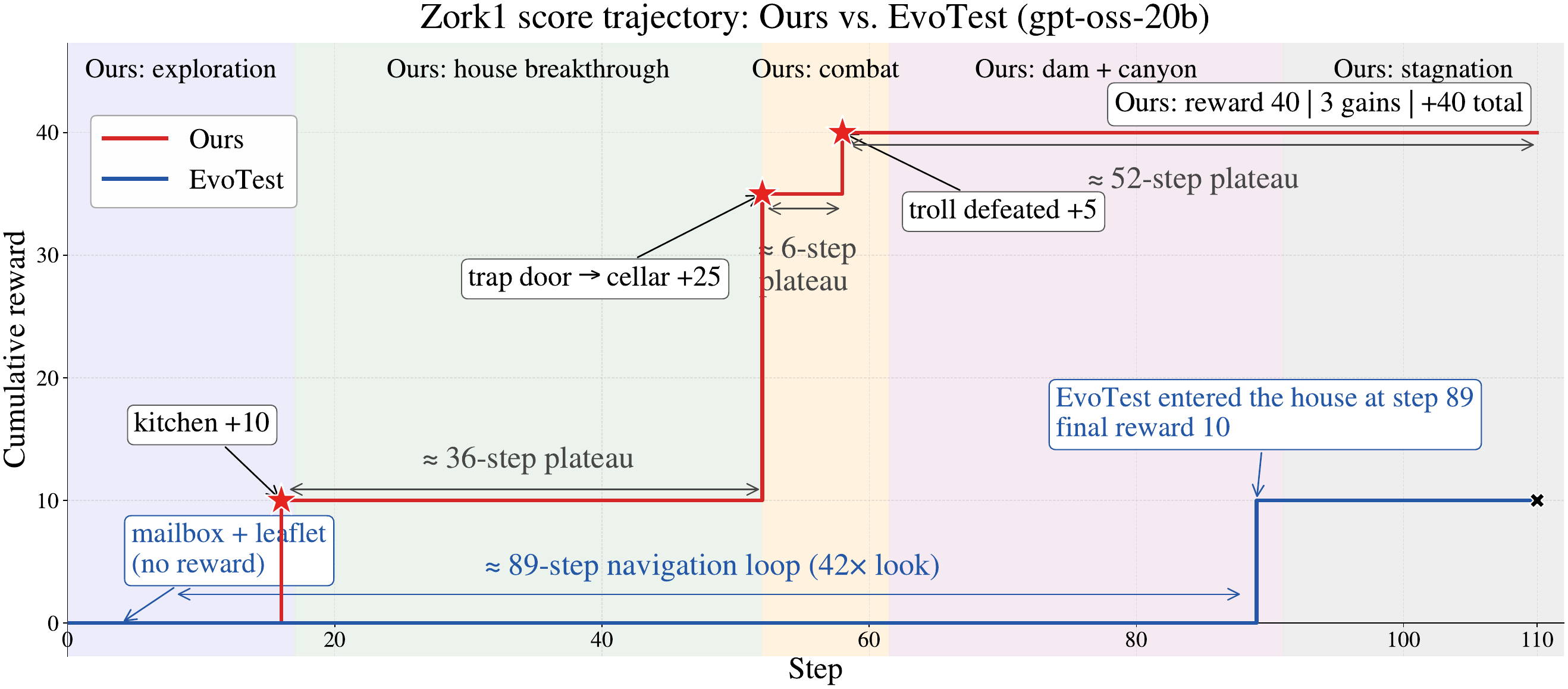}%
        \hspace{3.5mm}%
        \includegraphics[height=0.29\textwidth]
        {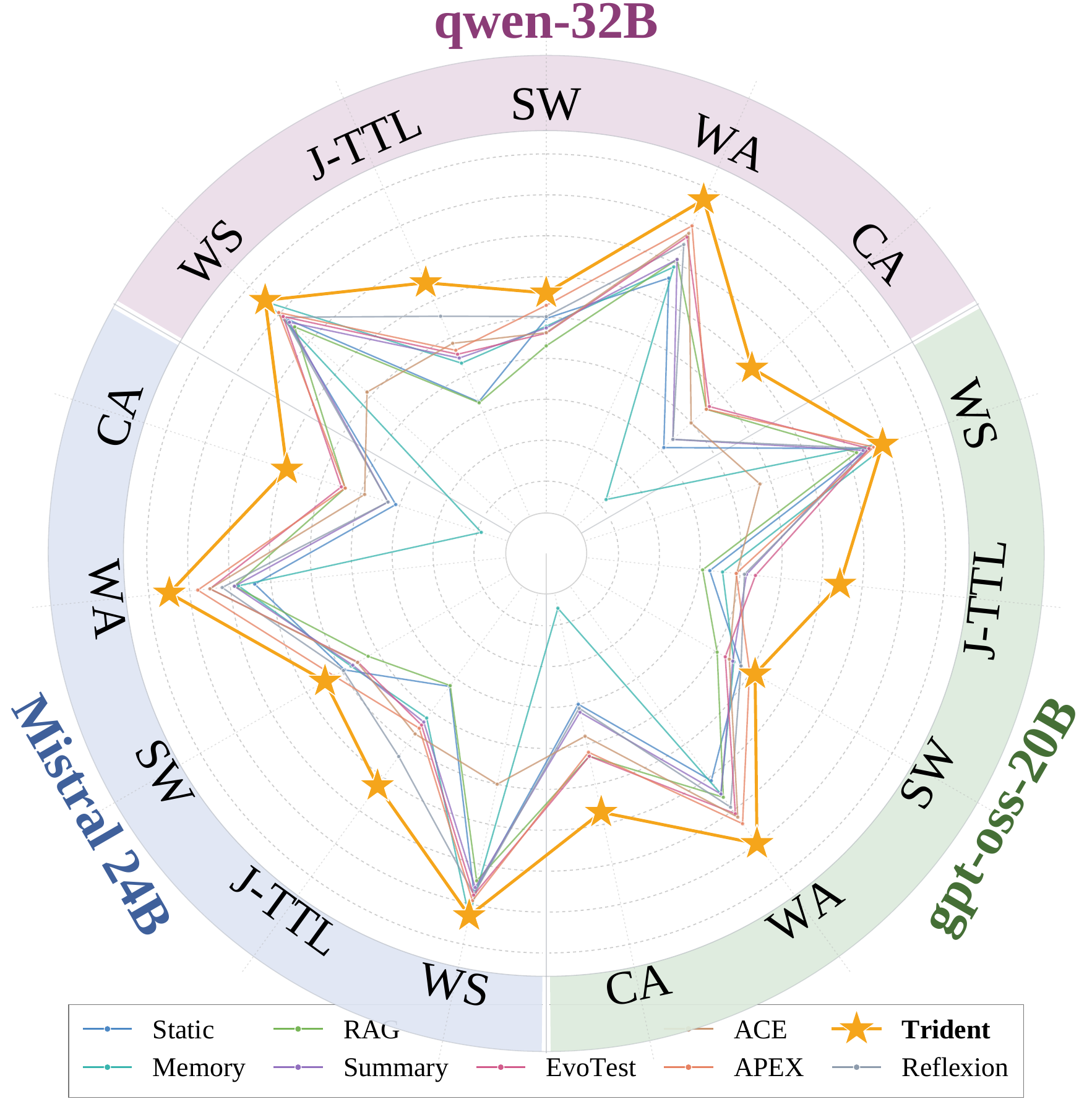}%
    }
    \caption{
        \textbf{Left:} \method achieves substantially higher reward than EvoTest on Zork1.
        \textbf{Right:} \method consistently outperforms baselines across tasks and backbones.
    }
    \label{fig:motivation}
\end{figure*}

\section{Introduction}
\label{sec:intro}
 
Large Language Model (LLM)-based agents
have been widely applied to web navigation
~\citep{zhou2024webarena},
embodied and text-based games
~\citep{wang2022scienceworld},
code generation
~\citep{fu2025catarena},
and tool-augmented problem solving
~\citep{shinn2023reflexion}.
External memory, reflection, planning, retrieval,
and tool-use mechanisms enable agents
to adapt their behavior using historical feedback
~\citep{zhao2024expel}.
This capability is particularly important
for long-horizon tasks,
where agents must reuse effective experience
across episodes,
avoid repeated failures,
and continually refine their strategies.

Accordingly, recent research has increasingly focused on
\textit{test-time learning} for LLM agents:
the backbone model remains frozen,
while performance is improved by updating
external memory, reflections, prompts,
or the agentic harness.
Existing approaches mainly include
memory-based methods that retrieve historical trajectories
~\citep{zhong2024memorybank},
reflection-based methods that summarize failures
and revise subsequent decisions
~\citep{zhang2024agent},
and evolutionary or search-based methods
that optimize prompts, policies,
and agentic workflows
~\citep{fernando2024promptbreeder,
khattab2024dspy}.
EvoTest~\citep{he2026evotest}
further formalizes this process as an
\textit{Act-Evolve} loop:
the Actor interacts with the environment
using the current harness,
and the Evolver updates the harness
from the completed trajectory,
enabling continual improvement
without gradient updates,
additional supervision,
or environment-specific fine-tuning.

However, despite its promise, trajectory-conditioned test-time evolution still suffers from several key information bottlenecks. First, deployment-time memory retrieval is often dominated by textual similarity, implicitly assuming that the most relevant memory is also the most useful one (\textbf{Dilemma 1})~\citep{ye2026umem, wang2026mem}. In partially observable environments, two states may look similar while differing in hidden prerequisites, inventory, recent actions, or task progress; directly injecting highly similar memories can therefore mislead the Actor or encourage shortcut copying. Second, the Evolver may generate multiple child harnesss that are lexically different but behaviorally redundant (\textbf{Dilemma 2})~\citep{ursekar2026vero}. If the downstream selector explores such near-duplicate harnesss, test-time computation is wasted on repeated strategies rather than genuinely complementary behaviors. Third, existing evolution procedures usually rewrite the whole harness in one shot (\textbf{Dilemma 3})~\citep{shen2025alignment}. Since an agentic harness contains coupled components, including policy prompt, success memory, failure memory, hyperparameters, and tool-use routines, monolithic rewriting may overwrite useful parent content or introduce inconsistency across components. These dilemmas reveal that improving test-time agent evolution requires not only stronger evolution, but also better \textit{memory reading}, \textit{harness selection}, and \textit{harness writing}.

To address these challenges, we propose \textbf{\method} (\textbf{T}est-time \textbf{R}obust \textbf{I}nterface-aware \textbf{D}iversified \textbf{E}volutio\textbf{N} for agen\textbf{T}s), which improves three information interfaces in the Act-Evolve loop. On the Actor side, \method separates memory relevance from utility and prioritizes boundary-relevant evidence that informs decisions without encouraging action copying. At the Evolver-to-Actor handoff, it selects task-relevant and behaviorally diverse harnesses before UCB routing. During harness evolution, it replaces whole-harness rewriting with ordered, slot-specific patches derived from trajectory events. Together, these designs improve experience utilization, behavioral exploration, and update stability while preserving EvoTest's gradient-free setting.

We conduct extensive experiments on five benchmarks across three domains: games (\textsc{J-TTL}, \textsc{ScienceWorld}), websites (\textsc{WebArena}, \textsc{WebShop}), and coding (\textsc{CatArena}). We evaluate \method with three open-source LLM backbones, and compare it with non-learning baselines, memory- and reflection-based methods, and automated prompt optimization methods. Empirical results show that \method is \textbf{\ding{182} high-performing}, achieving the best average performance across all three backbones; \textbf{\ding{183} robust across domains}, consistently improving on interactive games, web tasks, and coding environments; and \textbf{\ding{184} stable under parameter variation}, where the key hyperparameters exhibit clear non-monotonic trends and validate the design choices. Compared with the strongest baseline APEX, \method improves the average score by $0.0951$, $0.0914$, and $0.1007$ absolute points on \modelname{qwen-32B}, \modelname{mistral-small-3.2-24B}, and \modelname{gpt-oss-20B}, corresponding to relative gains of $30.7\%$, $30.8\%$, and $38.3\%$, respectively. Briefly put, our key contributions are summarized as follows:

\begin{itemize}[leftmargin=*,itemsep=-0.2em]
\item \textbf{\textit{Problem Diagnosis}:} We identify three information bottlenecks in trajectory-conditioned test-time agent evolution: similarity-biased memory injection, redundant child harness selection, and inconsistent whole-harness rewriting.
\item \textbf{\textit{Practical Framework}:}
We propose \method, a unified framework that improves how historical experience is retrieved, selected, and incorporated during test-time agent evolution, enabling more reliable and efficient adaptation without updating the backbone model.

\item \textbf{\textit{Experimental Evaluation}:} Extensive experiments on five benchmarks and three LLM backbones demonstrate that \method consistently outperforms non-learning, memory-based, reflection-based, and automated optimization baselines, with up to $38.3\%$ relative improvement over the strongest baseline.
\end{itemize}



 




 

\section{Related Work}

\noindent\textbf{Memory Retrieval and Utilization in Interactive Agents.}
LLM-based agents use episodic memory to reuse prior trajectories, including success/failure demonstrations~\cite{zhao2024expel}, long-term memories with recency or importance scores~\cite{zhong2024memorybank}, and hierarchical or graph-structured memories for long-horizon partially observable tasks~\cite{li2024optimus1,anokhin2025arigraph,sarch2024vlm}. Retrieval-augmented generation further shows that relevant context is not always useful, motivating critique, robustness filtering, and utility-aware reranking beyond similarity~\cite{asai2024selfrag,yoran2024making,yu2024rankrag,zhao2024seer}\cite{zou-etal-2026-uni,Su_2026_CVPR,DBLP:conf/www/SuZWXW26,yu2022deep,dou2024adversarial,zhao2026defending,dou2025toward,pan2026frequency,dou2026securesplit,wang2026cmphysbench,dou2026reshift,dou2026amrmpure,shui2026protoguard,li2025treble,tang2022few,liu2023spts,feng2024docpedia,zhao2024multi,wang2025pargo,tang2023character,lu2024bounding,zhao2025tabpedia,tang2024mtvqa,tang2024textsquare,shan2024mctbench,feng2023unidoc,tang2022youcan,fu2024ocrbench,wang2025wilddoc,feng2025dolphin,xiao2026staying,li2026aim}. \method reranks success, failure, and strategy memories by boundary-aware action utility, improving EvoTest retrieval under a fixed per-step budget.

\noindent\textbf{Configuration Selection and Optimization for Agentic Systems.}
Prompts, instructions, workflows, and agent designs can be viewed as searchable configurations. Prior work selects prompts through exploration--exploitation~\cite{shi2024efficient,wu2024ease,lin2024instinct,cai2025reinforcement}, mitigates behavioral redundancy among high-scoring candidates~\cite{wang2024effective,liu2026automated,liu2026toolanchor}, and searches over agent graphs, modular design spaces, or workflow operators~\cite{zhuge2024gptswarm,shang2025agentsquare}. Test-time optimization similarly allocates compute across candidates using process rewards, tree search, or verifier-guided selection~\cite{zhang2024rest,light2025strategist,snell2025scaling}\cite{wan2025srpo,dou2026dsadf,dou2026plan,lu2025prolonged,fei2025advancing,zhang2025hyperadalora,zhang2025trimtokenatorlc,zhang2025pdtrim,zhang2025trimtokenator,zhang2025sensitivity,mo2026shieldedcode,yu2026probability,zhang2026mitigating,li2026cpgrec+,li2025chi,li2024category,xiao2026layer,xiao2026not,xiao2026promptbased,xiao2025visual,jiang2024optimal,jiang2025thermochromic,tang2022optimal}. \method reduces behavioral redundancy among textually distinct EvoTest children by combining proxy-state relevance with configuration diversity, enabling UCB routing over a de-duplicated pool of prompt, memory, hyperparameter, and tool-use variants\cite{liu2026automated,liu2026toolanchor,zhao2025redone,shi2026androtmem,zhang2026temporal,yu2026cirrusbench,guo2025pet,zhang2026seizure,chen2026learning,ye2026look,chen2026understanding,ji2026tiny,zhang2026openhoi,huang2026skill,cai2025reinforcement,cai2026vi,yuan2026sp,kuang2025learning,nie2026understanding,wang2024muchin}.

\begin{figure*}[t]
\centering
\includegraphics[width= \textwidth]{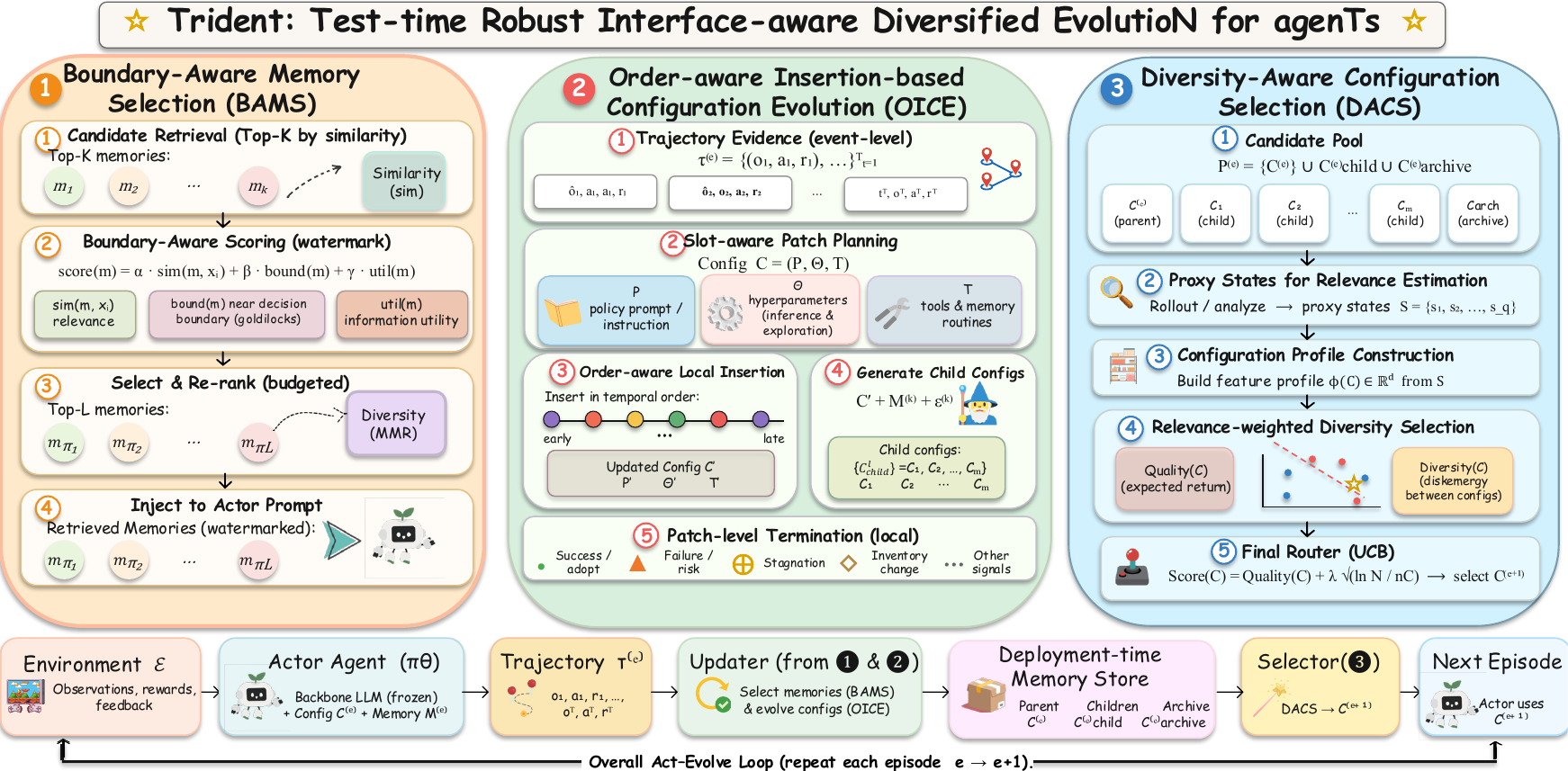}

\caption{Overview of \method}

\label{fig:main}

\end{figure*}

\noindent\textbf{Structured Generation and Evolution of Agent Configurations.}
Prior methods optimize prompts and agent configurations through full rewrites, population-level mutation, or free-form edits guided by execution feedback~\cite{guo2024evoprompt,fernando2024promptbreeder,yang2024opro,wang2024promptagent,khattab2024dspy,ma2024eureka,ijcai2024p3,qiao2024autoact,zhang2024agent,song2024trial,xiong2024watch,qu2024recursive,cheng2024trace,cai2026vi}\cite{ma2026thinking,macadmorph,Ma_Chen_Zhang_Wu_Ding_2025,10.1145/3731715.3733405,zou2026advancing,zou2026echogen,zou2026hiar,zheng2025vbench,pan2026eldiff,wen2026devil,zhao2024harmonizing,sun2025attentive,guo2025seed1,wang2025vision,zhao2026hieramp,wang2024muchin}. Such unconstrained updates may overwrite useful components or introduce cross-component inconsistency~\cite{ma2024death}. \method instead converts trajectory events into ordered, slot-specific patches with patch-level termination, enabling targeted updates while preserving effective parent structure.

\section{\method}\label{sec:method}
Following EvoTest~\citep{he2026evotest}, we redesign three key information interfaces, as illustrated in \Cref{fig:main}. Specifically, \method improves memory utilization before Actor decisions through boundary-aware memory selection, routes behaviorally diverse candidate harnesses to the Actor through diversity-aware harness selection, and incrementally incorporates trajectory evidence into structured harnesses through order-aware insertion-based harness evolution.

\subsection{Boundary-Aware Memory Selection(BAMS)}
\label{sec:bams}

Similar observations or abstract states may still differ in inventory, subgoals, prerequisites, or action history. BAMS therefore reranks the candidate memories returned by the inherited retriever according to relevance, historical usefulness, and boundary value, selecting non-conflicting evidence that informs decisions without encouraging rote action copying.

At decision step $t$ in episode $e$, BAMS first uses the memory retriever to obtain candidate memories $\mathcal{M}_{t}^{\mathrm{cand}}=\mathcal{R}_{M}(o_t^{(e)},s_t^{(e)},M^{(e)})=\mathcal{M}_{t}^{+}\cup\mathcal{M}_{t}^{-}$, where $\mathcal{M}_{t}^{+}$ and $\mathcal{M}_{t}^{-}$ denote recalled success and failure memories. Each candidate memory $m_i$ is scored by relevance, empirical usefulness, and boundary value:
\begin{equation}
\label{eq:bams_score}
\begin{aligned}
S_{\mathrm{BAMS}}(m_i; o_t, s_t)
=&\ \lambda_r \mathrm{Rel}(m_i,o_t,s_t)
+\lambda_u \mathrm{Use}(m_i) \\
&+\lambda_b \mathrm{Bnd}(m_i,o_t,s_t),
\end{aligned}
\end{equation}
where $\lambda_r,\lambda_u,\lambda_b\geq0$ and $\lambda_r+\lambda_u+\lambda_b=1$.

The relevance term combines semantic similarity and structured state overlap: $\mathrm{Rel}(m_i,o_t,s_t)=\alpha \overline{\cos}(E(m_i),E(o_t,s_t))+(1-\alpha)\mathrm{Jac}(\phi(m_i),\phi(o_t,s_t))$, where $\overline{\cos}=(\cos+1)/2$ normalizes cosine similarity to $[0,1]$, and $\phi(\cdot)$ extracts symbolic slots such as location, inventory, visible objects, exits, completed subgoals, and recent actions. The usefulness term is estimated from previous retrievals as $\mathrm{Use}(m_i)=\operatorname{clip}_{[0,1]}((n_i+\epsilon)^{-1}\sum_{\ell=1}^{n_i}(\Delta R_{i,\ell}/R_{\max}+\beta A_{i,\ell}))$, where $\Delta R_{i,\ell}$ is the short-horizon reward gain after using $m_i$, and $A_{i,\ell}=1$ if a failure memory helps avoid a repeated invalid or harmful action. Memories without retrieval history receive a neutral usefulness score.

The boundary term favors memories that are informative but not trivially identical to the current state:
\begin{equation}
\label{eq:bams_boundary}
\mathrm{Bnd}(m_i,o_t,s_t)
=
\exp\left(
-\frac{\bigl(\mathrm{Rel}(m_i,o_t,s_t)-c\bigr)^2}{2\sigma^2}
\right).
\end{equation}
Here, $c$ specifies the target relevance level and $\sigma$ controls the tolerance around it. We use $c=0.5$ by default and tune $\sigma$ on a small validation set. This operationalizes boundary awareness as intermediate relevance: low-relevance memories are usually irrelevant, whereas near-duplicate memories often encourage brittle action copying. Memories around $c$ are close enough to be useful but still require state-conditioned reasoning.

To avoid prompt pollution, BAMS selects a compact and non-conflicting memory subset:
\begin{equation}
\label{eq:bams_budget}
\resizebox{\columnwidth}{!}{$
\displaystyle
\hat{\mathcal{M}}_t
=
\arg\max_{\mathcal{S}\subseteq\mathcal{M}_{t}^{\mathrm{cand}},\,|\mathcal{S}|\le B}
\left[
\sum_{m_i\in\mathcal{S}}
S_{\mathrm{BAMS}}(m_i;o_t,s_t)
-
\eta
\sum_{m_i,m_j\in\mathcal{S}}
\mathrm{Conf}(m_i,m_j)
\right]
$}
\end{equation}
The conflict score penalizes memories that match similar states but suggest incompatible outcomes or actions: $\mathrm{Conf}(m_i,m_j)=\mathbf{1}\{\mathrm{Rel}(m_i,m_j)>\tau_c\}\cdot\mathbf{1}\{y_i\neq y_j\ \mathrm{or}\ a_i\not\sim a_j\}$, where $y_i$ is the success/failure label and $a_i\not\sim a_j$ denotes incompatible action slots, such as different verbs or mutually exclusive objects under the same state condition. Although Eq.~\ref{eq:bams_budget} is written as a subset objective, we solve it greedily by repeatedly adding the memory with the largest positive marginal score $S_{\mathrm{BAMS}}(m_i;o_t,s_t)-\eta\sum_{m_j\in\mathcal{S}}\mathrm{Conf}(m_i,m_j)$ until the budget $B$ is reached.


\begin{table*}[t]
\caption{
Overall performance comparison across five benchmarks.
All methods are evaluated with the same backbone model within each block.
We highlight the \hlfirst{best} and \hlsecond{second best} results.
All methods follow the same evaluation protocol and budget constraints described below.
$^{\dagger}$: No-learning Baseline.
$^{\star}$: Memory-Based \& Reflection-based Methods.
$^{\diamond}$: Automated Prompt Optimization Methods.
}
\label{tab:main}
\centering
\footnotesize
\setlength{\tabcolsep}{3pt}
\resizebox{0.95\textwidth}{!}{%
\begin{tabular}{@{}c|l|cccccc@{}}
\Xhline{1.2pt}
\rowcolor{CadetBlue!20}
\textbf{Backbone}
& \textbf{Method}
& \textbf{WebShop}
& \textbf{J-TTL}
& \textbf{ScienceWorld}
& \textbf{WebArena}
& \textbf{CatArena}
& \textbf{Avg.} \\
\Xhline{1.2pt}

\multirow{9}{*}{\rotatebox[origin=c]{90}{\modelname{qwen-32B}}}
& \methodname{Static$^{\dagger}$}
& 0.4823\red{0.0000}
& 0.0290\red{0.0000}
& 0.1976\red{0.0000}
& 0.3600\red{0.0000}
& 0.0100\red{0.0000}
& 0.2158\red{0.0000} \\

& \methodname{Memory$^{\star}$}
& \hlsecond{0.5403\red{0.0580}}
& 0.1324\red{0.1034}
& 0.1786\blue{0.0190}
& 0.3900\red{0.0300}
& -0.1800\blue{0.1900}
& 0.2123\blue{0.0035} \\

& \methodname{RAG$^{\star}$}
& 0.4510\blue{0.0313}
& 0.0263\blue{0.0027}
& 0.1307\blue{0.0669}
& 0.4100\red{0.0500}
& 0.1500\red{0.1400}
& 0.2336\red{0.0178} \\

& \methodname{Summary$^{\star}$}
& 0.4680\blue{0.0143}
& 0.1461\red{0.1171}
& 0.1745\blue{0.0231}
& 0.4100\red{0.0500}
& 0.0400\red{0.0300}
& 0.2477\red{0.0319} \\

& \methodname{Reflexion$^{\star}$}
& 0.4859\red{0.0036}
& \hlsecond{0.2580\red{0.2290}}
& 0.2023\red{0.0047}
& 0.4500\red{0.0900}
& 0.0400\red{0.0300}
& 0.2872\red{0.0714} \\

& \methodname{EvoTest$^{\diamond}$}
& 0.4889\red{0.0066}
& 0.1562\red{0.1272}
& 0.1616\blue{0.0360}
& 0.4700\red{0.1100}
& \hlsecond{0.1600\red{0.1500}}
& 0.2873\red{0.0715} \\

& \methodname{ACE$^{\diamond}$}
& 0.2128\blue{0.2695}
& 0.1855\red{0.1565}
& 0.1630\blue{0.0346}
& 0.4800\red{0.1200}
& 0.1000\red{0.0900}
& 0.2283\red{0.0125} \\

& \methodname{APEX$^{\diamond}$}
& 0.5034\red{0.0211}
& 0.1668\red{0.1378}
& \hlsecond{0.2297\red{0.0321}}
& \hlsecond{0.5000\red{0.1400}}
& 0.1500\red{0.1400}
& \hlsecond{0.3100\red{0.0942}} \\

& \method
& \hlfirst{0.5480\red{0.0657}}
& \hlfirst{0.3476\red{0.3186}}
& \hlfirst{0.2600\red{0.0624}}
& \hlfirst{0.5700\red{0.2100}}
& \hlfirst{0.3000\red{0.2900}}
& \hlfirst{0.4051\red{0.1893}} \\

\midrule

\multirow{9}{*}{\rotatebox[origin=c]{90}{\modelname{mistral-small-3.2-24B}}}
& \methodname{Static$^{\dagger}$}
& 0.4727\red{0.0000}
& 0.0251\red{0.0000}
& 0.1926\red{0.0000}
& 0.3400\red{0.0000}
& 0.0100\red{0.0000}
& 0.2081\red{0.0000} \\

& \methodname{Memory$^{\star}$}
& \hlfirst{0.5313\red{0.0586}}
& 0.1205\red{0.0954}
& 0.1741\blue{0.0185}
& 0.3800\red{0.0400}
& -0.2100\blue{0.2200}
& 0.1992\blue{0.0089} \\

& \methodname{RAG$^{\star}$}
& 0.4414\blue{0.0313}
& 0.0225\blue{0.0026}
& 0.1260\blue{0.0666}
& 0.3900\red{0.0500}
& 0.1400\red{0.1300}
& 0.2240\red{0.0159} \\

& \methodname{Summary$^{\star}$}
& 0.4584\blue{0.0143}
& 0.1327\red{0.1076}
& 0.1695\blue{0.0231}
& 0.3900\red{0.0500}
& 0.0300\red{0.0200}
& 0.2361\red{0.0280} \\

& \methodname{Reflexion$^{\star}$}
& 0.4753\red{0.0026}
& \hlsecond{0.2368\red{0.2117}}
& 0.1959\red{0.0033}
& 0.4200\red{0.0800}
& 0.0300\red{0.0200}
& 0.2716\red{0.0635} \\

& \methodname{EvoTest$^{\diamond}$}
& 0.4774\red{0.0047}
& 0.1419\red{0.1168}
& 0.1545\blue{0.0381}
& 0.4500\red{0.1100}
& \hlsecond{0.1500\red{0.1400}}
& 0.2748\red{0.0667} \\

& \methodname{ACE$^{\diamond}$}
& 0.2000\blue{0.2727}
& 0.1686\red{0.1435}
& 0.1560\blue{0.0366}
& 0.4500\red{0.1100}
& 0.0900\red{0.0800}
& 0.2129\red{0.0048} \\

& \methodname{APEX$^{\diamond}$}
& 0.4912\red{0.0185}
& 0.1514\red{0.1263}
& \hlsecond{0.2221\red{0.0295}}
& \hlsecond{0.4800\red{0.1400}}
& 0.1400\red{0.1300}
& \hlsecond{0.2969\red{0.0888}} \\

& \method
& \hlsecond{0.5288\red{0.0561}}
& \hlfirst{0.3249\red{0.2998}}
& \hlfirst{0.2478\red{0.0552}}
& \hlfirst{0.5500\red{0.2100}}
& \hlfirst{0.2900\red{0.2800}}
& \hlfirst{0.3883\red{0.1802}} \\

\midrule

\multirow{9}{*}{\rotatebox[origin=c]{90}{\modelname{gpt-oss-20B}}}
& \methodname{Static$^{\dagger}$}
& 0.4523\red{0.0000}
& 0.0253\red{0.0000}
& 0.1731\red{0.0000}
& 0.3100\red{0.0000}
& 0.0000\red{0.0000}
& 0.1921\red{0.0000} \\

& \methodname{Memory$^{\star}$}
& \hlfirst{0.5123\red{0.0600}}
& 0.0567\red{0.0314}
& 0.1530\blue{0.0201}
& 0.3500\red{0.0400}
& -0.2400\blue{0.2400}
& 0.1664\blue{0.0257} \\

& \methodname{RAG$^{\star}$}
& 0.4210\blue{0.0313}
& 0.0076\blue{0.0177}
& 0.1058\blue{0.0673}
& 0.3600\red{0.0500}
& \hlsecond{0.1300\red{0.1300}}
& 0.2049\red{0.0128} \\

& \methodname{Summary$^{\star}$}
& 0.4380\blue{0.0143}
& 0.1155\red{0.0902}
& 0.1497\blue{0.0234}
& 0.3500\red{0.0400}
& 0.0200\red{0.0200}
& 0.2146\red{0.0225} \\

& \methodname{Reflexion$^{\star}$}
& 0.4529\red{0.0006}
& 0.1105\red{0.0852}
& 0.1732\red{0.0001}
& 0.3900\red{0.0800}
& 0.0100\red{0.0100}
& 0.2273\red{0.0352} \\

& \methodname{EvoTest$^{\diamond}$}
& 0.4529\red{0.0006}
& \hlsecond{0.1376\red{0.1123}}
& 0.1287\blue{0.0444}
& 0.4100\red{0.1000}
& \hlsecond{0.1300\red{0.1300}}
& 0.2518\red{0.0597} \\

& \methodname{ACE$^{\diamond}$}
& 0.1728\blue{0.2795}
& 0.0916\red{0.0663}
& 0.1401\blue{0.0330}
& 0.4200\red{0.1100}
& 0.0800\red{0.0800}
& 0.1809\blue{0.0112} \\

& \methodname{APEX$^{\diamond}$}
& 0.4654\red{0.0131}
& 0.0903\red{0.0650}
& \hlsecond{0.1972\red{0.0241}}
& \hlsecond{0.4400\red{0.1300}}
& 0.1200\red{0.1200}
& \hlsecond{0.2626\red{0.0705}} \\

& \method
& \hlsecond{0.4880\red{0.0357}}
& \hlfirst{0.3453\red{0.3200}}
& \hlfirst{0.2130\red{0.0399}}
& \hlfirst{0.5000\red{0.1900}}
& \hlfirst{0.2700\red{0.2700}}
& \hlfirst{0.3633\red{0.1712}} \\

\Xhline{1.2pt}
\end{tabular}%
}
\end{table*}

\subsection{Diversity-Aware Harness Selection(DAHS)}
\label{sec:DAHS}
After each episode, the Evolver may generate multiple child harnesses, many of which differ lexically but induce similar action routines, memory usage, or failure patterns. To avoid wasting UCB exploration on near-duplicates, we profile candidates by their configurations and proxy-state behaviors, then select a relevance-aware and behaviorally diverse subset for routing to the next Actor.

For each candidate harness $C_i\in\mathcal{P}^{(e)}$, DAHS builds a harness profile $\Phi(C_i)=[\phi(P_i),\phi(\Theta_i),\phi(\mathcal{T}_i),\phi(\mathcal{B}_i)]$, where $P_i$ denotes prompt rules, $\Theta_i$ denotes decoding and retrieval hyperparameters, $\mathcal{T}_i$ denotes tool-use routines, and $\mathcal{B}_i$ denotes proxy-state action traces under the shared memory store $M^{(e)}$. Each component is encoded as a normalized feature vector, allowing DAHS to capture behavioral similarity beyond prompt-level overlap.

DAHS estimates candidate relevance on proxy states extracted from historical trajectories, denoted by $\mathcal{S}_{\mathrm{proxy}}^{(e)}=\mathcal{S}_{+}^{(e)}\cup\mathcal{S}_{-}^{(e)}\cup\mathcal{S}_{\mathrm{loop}}^{(e)}\cup\mathcal{S}_{\mathrm{prog}}^{(e)}$. Here, $\mathcal{S}_{+}^{(e)}$ contains states preceding reward gains, $\mathcal{S}_{-}^{(e)}$ contains invalid or dead-end states, $\mathcal{S}_{\mathrm{loop}}^{(e)}$ contains repeated state-action patterns, and $\mathcal{S}_{\mathrm{prog}}^{(e)}$ contains intermediate transitions that changed the abstract state. For each proxy state $s_j$, DAHS queries the candidate harness once to obtain a proposed action $a_i(s_j)$.

Candidate relevance is computed as $\mathrm{Rel}_{C}(C_i)=|\mathcal{S}_{\mathrm{proxy}}^{(e)}|^{-1}\sum_{s_j\in\mathcal{S}_{\mathrm{proxy}}^{(e)}}\mathcal{R}_{\mathrm{cfg}}(C_i,s_j)$, where $\mathcal{R}_{\mathrm{cfg}}(C_i,s_j)=I_{+}(C_i,s_j)-I_{\mathrm{bad}}(C_i,s_j)$. Here, $I_{+}(C_i,s_j)=1$ if $a_i(s_j)$ matches a historically useful action for $s_j$, and $0$ otherwise. Similarly, $I_{\mathrm{bad}}(C_i,s_j)=1$ if $a_i(s_j)$ matches an action that previously led to failure or loops, and $0$ otherwise. The risk set is $\mathcal{S}_{\mathrm{risk}}^{(e)}=\mathcal{S}_{-}^{(e)}\cup\mathcal{S}_{\mathrm{loop}}^{(e)}$, and the risk score is $\mathrm{Risk}(C_i)=|\mathcal{S}_{\mathrm{risk}}^{(e)}|^{-1}\sum_{s_j\in\mathcal{S}_{\mathrm{risk}}^{(e)}}I_{\mathrm{bad}}(C_i,s_j)$.

DAHS first selects a diversified subset by maximizing a relevance-risk-diversity objective:
\begin{equation}
\label{eq:DAHS_subset_score}
\resizebox{\columnwidth}{!}{$
\displaystyle
\hat{\mathcal{P}}^{(e)}
=
\arg\max_{\mathcal{S}\subseteq\mathcal{P}^{(e)},\,|\mathcal{S}|=K}
\left[
\sum_{C_i\in\mathcal{S}}
\left(
\alpha \mathrm{Rel}_{C}(C_i)
-
\gamma \mathrm{Risk}(C_i)
\right)
+
\beta \log\det(\mathbf{K}_{\mathcal{S}})
\right]
$}
\end{equation}
The kernel $\mathbf{K}_{\mathcal{S}}$ is built from harness profiles, with $\mathbf{K}_{ij}=\exp(-\|\Phi(C_i)-\Phi(C_j)\|_2^2/\rho)+\epsilon I_{ij}$, where $I_{ij}=1$ if $i=j$ and $0$ otherwise. The log-determinant term encourages complementary strategies instead of redundant prompt variants. In practice, DAHS greedily adds the candidate with the largest marginal increase in the objective until $K$ harnesses are selected.

Finally, DAHS selects the next harness from the diversified pool by augmenting the inherited UCB criterion with candidate relevance and risk:
\begin{equation}
\label{eq:DAHS_ucb}
\resizebox{\columnwidth}{!}{$
C^{(e+1)}
=
\arg\max_{C_i\in\hat{\mathcal{P}}^{(e)}}
\left[
\bar{R}(C_i)
+\lambda \sqrt{\tfrac{\log(1+n_e)}{1+n_i}}
+\mu \mathrm{Rel}_{C}(C_i)
-\nu \mathrm{Risk}(C_i)
\right]
$}
\end{equation}
Thus, DAHS preserves EvoTest's exploration mechanism while ensuring that exploration is spent on behaviorally distinct and task-relevant harnesses.

\subsection{Order-Aware Insertion-Based Harness Evolution(OICE)}
\label{sec:oice}

OICE maps trajectory events to editable slots and updates the harness through ordered local patches, preserving effective structure while enabling targeted revisions.

Given a completed trajectory $\tau^{(e)}$, OICE extracts learning-relevant events $\mathcal{Z}^{(e)}=\mathcal{P}_{\mathrm{event}}(\tau^{(e)})=\{z_1,\dots,z_L\}$. Events are extracted by rule-based detectors over trajectory logs, including reward changes, repeated state-action pairs, invalid feedback, object-interaction feedback, inventory updates, and prerequisite-related failure messages. Each event is represented as $z_l=(x_l,a_l,y_l,\Delta r_l,\omega_l)$, where $x_l$ is the local context, $a_l$ is the action, $y_l$ is the feedback, $\Delta r_l$ is the reward change, and $\omega_l\in\{\mathrm{prog},\mathrm{invalid},\mathrm{loop},\mathrm{object},\mathrm{inv},\mathrm{prereq}\}$ is the event type.

OICE treats the harness as editable slots $C=(C_{\mathrm{prompt}},C_{\mathrm{hyp}},C_{\mathrm{tool}})$ and maps each event to one or more slots by $\mathcal{L}_l=\mathcal{L}_{\mathrm{slot}}(z_l,C)\subseteq\{\mathrm{prompt},\mathrm{hyp},\mathrm{tool}\}$. Progress, object, inventory, and prerequisite events are mapped to prompt rules; repeated invalid or loop events are mapped to prompt or hyperparameter slots; and tool-related failures are mapped to tool-use slots.

At insertion step $k$, OICE generates a local patch from the selected event-slot pair and inserts it into the current partial harness:
\begin{equation}
\label{eq:oice_patch_insert}
\resizebox{\columnwidth}{!}{$
\displaystyle
p_k
=
\mathcal{H}_{\mathrm{patch}}(z_{\pi_k},\ell_{\pi_k},C_{k-1}^{(e)}),
\quad
C_k^{(e)}
=
\mathcal{U}_{\mathrm{insert}}(C_{k-1}^{(e)},p_k,\ell_{\pi_k})
$}
\end{equation}
Here, a patch is a typed edit, such as adding a prompt rule, adjusting a retrieval or action budget, or inserting a tool-use constraint. The insertion operator only modifies the target slot and rejects a patch if it contradicts a higher-confidence existing rule.

The insertion order is trajectory-dependent:
\begin{equation}
\label{eq:oice_order}
\resizebox{\columnwidth}{!}{$
\displaystyle
\pi
=
\arg\max_{\pi'}
\sum_{k=1}^{K}
\left[
\mathrm{Gain}(z_{\pi'_k},\ell_{\pi'_k},C_{k-1}^{(e)})
-
\zeta
\mathrm{Conflict}(p_k,C_{k-1}^{(e)})
\right]
$}
\end{equation}
The gain score is computed as $\mathrm{Gain}(z_l,\ell,C)=|\Delta r_l|/R_{\max}+\lambda_f f_l+\lambda_s q_l$, where $f_l$ is the frequency of the same event pattern in the trajectory and $q_l$ indicates success evidence, missing prerequisites, or loop reduction. The conflict score is $\mathrm{Conflict}(p,C)=I_{\mathrm{same}}(p,C)+I_{\mathrm{contra}}(p,C)$, where $I_{\mathrm{same}}$ indicates redundant edits to the same slot and $I_{\mathrm{contra}}$ indicates contradiction with an existing rule or parameter constraint.

\section{Experiments}
\label{sec:experiments}

In this section, we conduct extensive experiments to answer: (RQ1) How does \method perform
compared to existing methods? (RQ2) How does each module contribute to \method's performance? (RQ3) What does \method actually discover, and why does it outperform baselines? (RQ4) How sensitive is \method to its key parameters?

\begin{figure*}[htbp]
    \centering
    \setlength{\tabcolsep}{1pt}
    \begin{tabular}{ccc}
        \includegraphics[width=0.325\textwidth]{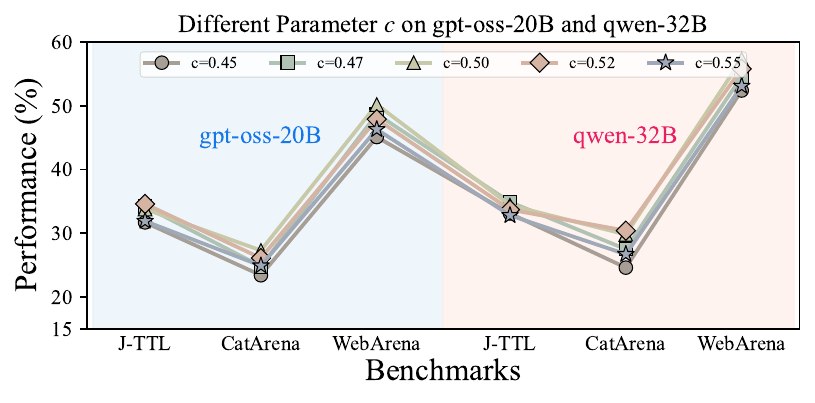}
        &
        \includegraphics[width=0.325\textwidth]{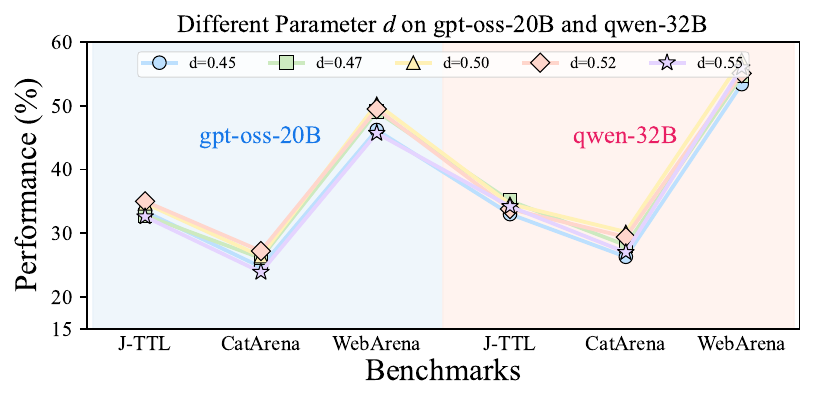}
        &
        \includegraphics[width=0.325\textwidth]{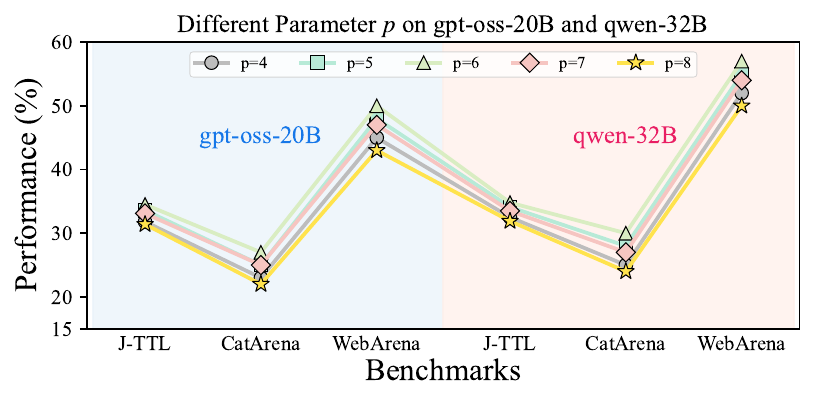}
     
    \end{tabular}
  
    \caption{
Sensitivity analysis of the memory relevance center $c$, diversity weight $\beta$, and maximum number of ordered patches per episode.
    }
    \label{fig:parameter_sensitivity}

\end{figure*}

\subsection{Experiment Setup}

\paragraph{Datasets and Benchmarks.}
We evaluate \method on five benchmarks across three domains: 
J-TTL~\cite{he2026evotest} and ScienceWorld~\cite{wang2022scienceworld} for games, 
WebArena~\cite{zhou2024webarena} and WebShop~\cite{yao2022webshop} for web tasks, 
and CatArena~\cite{fu2025catarena} for coding. 
We use the full J-TTL and CatArena benchmarks, all 812 WebArena tasks, and fixed 100-task subsets for ScienceWorld and WebShop. 
All methods use the same task sets and budgets: 10 episodes per task for J-TTL, CatArena, and WebShop, and 5 episodes per task for ScienceWorld and WebArena, with a per-episode step limit of 30 for WebShop and 120 otherwise.




\begin{figure}[t]
    \centering
    \includegraphics[width=1\columnwidth]{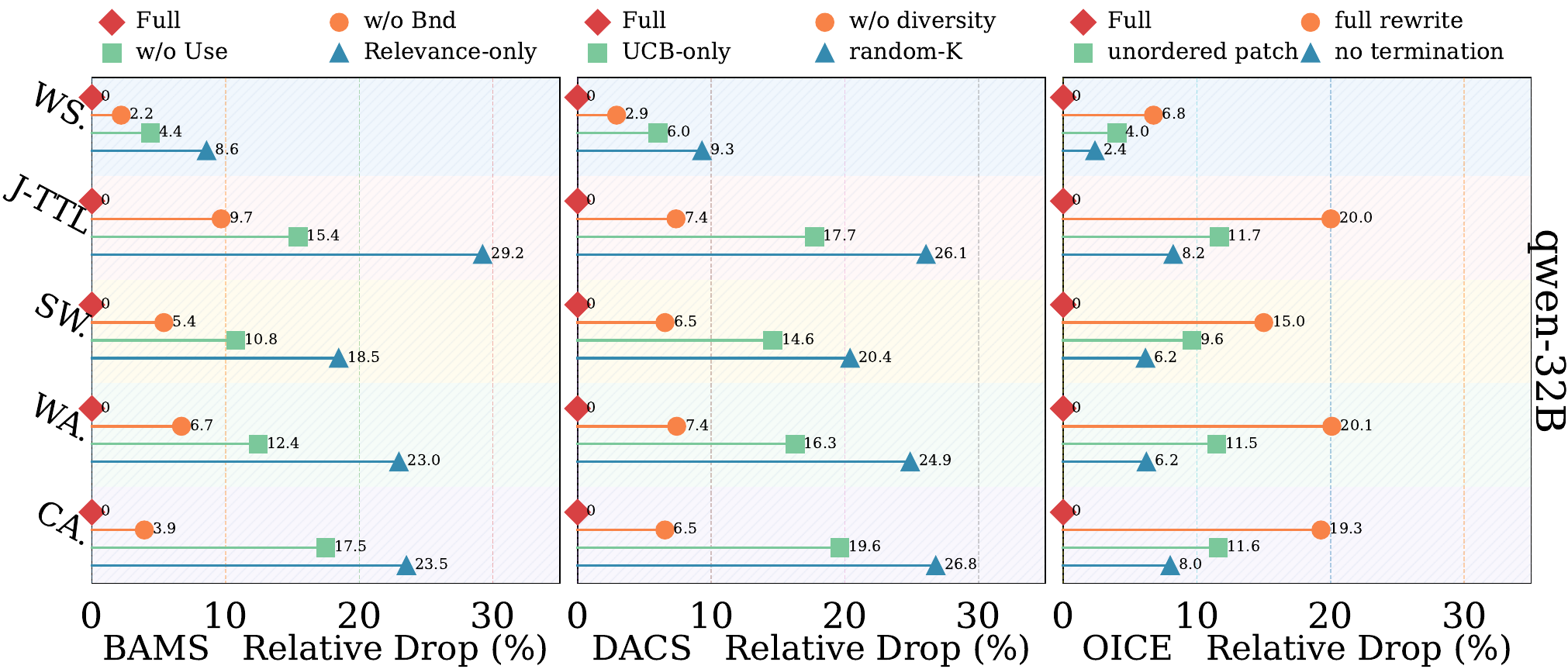}
    \caption{Mechanism-level ablation study of \method on \modelname{qwen-32B}}
    \label{fig:mechanism_ablation}
\end{figure}

\begin{figure}[t]
    \centering
    \includegraphics[width=1\columnwidth]{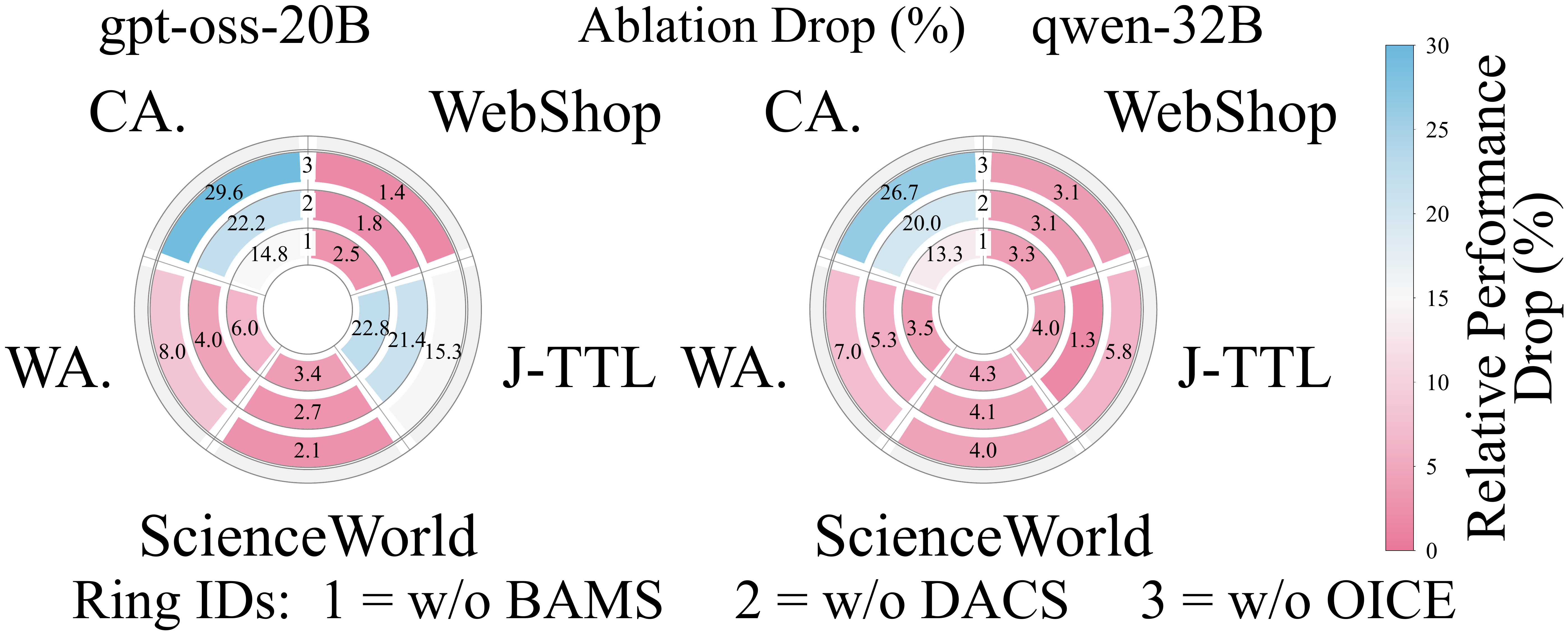}
    \caption{Ablation study of BAMS, DAHS, OICE.}
    \label{fig:ablation}
\end{figure}



















\begin{figure}[t]
    \centering
    \includegraphics[width=\linewidth]
    {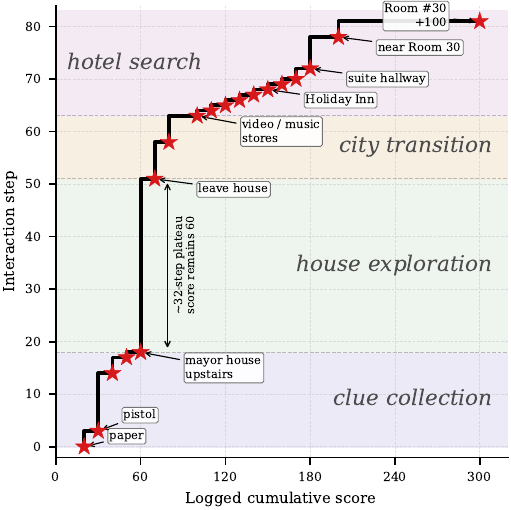}
    \caption{
        Case study on \textit{Detective} using
        \modelname{gpt-oss-20B}.
        The trajectory shows the highest-score discovery process.
    }
    \label{fig:case_detective}
\end{figure}

\paragraph{Baselines.}
We compare \method against eight baselines grouped into three categories:
\textbf{(1) No-learning baseline:}
\modelname{Static};
\textbf{(2) Memory- and reflection-based methods:}
\modelname{Memory}~\cite{he2026evotest},
\modelname{RAG}~\cite{he2026evotest},
\modelname{Summary}~\cite{he2026evotest}, and
\modelname{Reflexion}~\cite{shinn2023reflexion};
and \textbf{(3) Automated prompt optimization methods:}
\modelname{EvoTest}~\cite{he2026evotest},
\modelname{ACE}~\cite{zhang2025agentic}, and
\modelname{APEX}~\cite{li2026apex}.

\paragraph{LLM backbones.}For instantiating these baselines, we adopt three open-source LLMs, \modelname{qwen-32B}, \modelname{mistral-small-3.2-24B}, and \modelname{gpt-oss-20B}. 

\paragraph{Metric.}
We use benchmark-specific evaluation metrics. For WebShop, J-TTL, ScienceWorld, and WebArena, we report normalized Area Under the Curve (AUC)~\cite{he2026evotest}, defined as $\mathrm{AUC}=\frac{1}{K R_{\max}}\sum_{e=1}^{K}R(e)$, where $R(e)$ is the score obtained in episode $e$, $K$ is the number of episodes, and $R_{\max}$ is the maximum attainable score. The normalized AUC lies in $[0,1]$ and enables comparison across tasks with different score scales. For CatArena, we follow its official evaluation protocol~\cite{fu2025catarena} and report the evolutionary capability score $S_{\mathrm{evo}}$. For agent $i$ at round $n$, CatArena first computes its global performance $G_i^n$ by averaging the performance of its round-$n$ code against all other agents and code versions across all rounds; for objective tasks, $G_i^n$ is simply the performance score at round $n$. It then fits $G_i^n \approx k_i n+b_i$ using ordinary least squares and defines
$S_{\mathrm{evo},i}=k_i=\frac{\sum_{n=1}^{N}(n-\bar{n})(G_i^n-\bar{G}_i)}{\sum_{n=1}^{N}(n-\bar{n})^2}$,
where $\bar{n}$ and $\bar{G}_i$ denote the average round index and average global performance, respectively. A positive $S_{\mathrm{evo}}$ indicates sustained improvement across rounds, whereas a negative value indicates performance regression. Higher values are better for both metrics. The Avg. column in Table~\ref{tab:main} reports the unweighted macro-average of the five benchmark-specific metrics.

\paragraph{Parameter Configuration.}

For our proposed modules, most hyperparameters are fixed throughout the experiments. In BAMS, we consider at most $8$ candidate memories per decision step and inject up to $2$ successful memories, $1$ failure memory, and $1$ strategy-level hint. In DAHS, we first keep the top-$8$ candidate configurations and retain $4$ diversified candidates for the final UCB-based routing. The relevance, risk-penalty, UCB-bonus, and novelty weights are fixed to $1.0$, $0.7$, $0.75$, and $0.35$, respectively. In OICE, we extract at most $12$ trajectory events per episode. These parameters are not tuned per benchmark or per backbone.


\subsection{Main Results (RQ1)}
\paragraph{Takeaway~$1$: \method achieves the best overall performance among the compared methods across diverse task domains.}
As shown in Table~\ref{tab:main}, \method delivers the strongest average results across all evaluated backbone models. Compared with the strongest baseline \methodname{APEX}, \method improves the average score by $33.3\%$ on average, with backbone-specific relative gains of $30.7\%$, $30.8\%$, and $38.3\%$, respectively. Moreover, \method ranks first in $13$ of the $15$ benchmark--backbone settings, including all settings on J-TTL, ScienceWorld, WebArena, and CatArena. These results demonstrate that \method provides robust and generalizable benefits across heterogeneous benchmarks and backbone models.

\subsection{Ablation Study (RQ2)}
\paragraph{Takeaway~$2$: All three modules contribute consistently, with OICE providing the largest average gain.}
As shown in Figure~\ref{fig:ablation}, removing BAMS, DAHS, or OICE consistently degrades performance. On \modelname{gpt-oss-20B}, the relative drops are $9.28\%$, $9.30\%$, and $10.16\%$, respectively; on \modelname{qwen-32B}, they are $5.09\%$, $6.02\%$, and $8.27\%$. Averaged across both backbones, the drops are $7.19\%$, $7.66\%$, and $9.22\%$, with OICE having the largest impact, highlighting the importance of ordered local updates for stable evolution.

\paragraph{Takeaway~$3$: Each module's internal design is necessary for reliable retrieval, diverse routing, and stable updating.}
Figure~\ref{fig:mechanism_ablation} further validates the internal design of each module. Averaged over WebShop, J-TTL, and ScienceWorld, relevance-only retrieval in BAMS causes a $17.01\%$ drop, showing that similarity alone cannot identify useful memories. For DAHS, random selection, UCB-only routing, and removing diversity reduce performance by $16.84\%$, $11.48\%$, and $5.07\%$, respectively, confirming the need to jointly consider relevance, diversity, and risk. For OICE, full rewriting, unordered patching, and removing termination cause drops of $12.59\%$, $7.58\%$, and $4.98\%$, respectively, demonstrating that controlled ordered updates better preserve effective behaviors.

 \subsection{Analysis: What Does \method Discover? (RQ3)}
\label{sec:case_study}

\paragraph{Takeaway~$4$: \method extracts milestone-level strategy chains and exposes exploration bottlenecks.}
We answer \textbf{RQ3} by analyzing the strategy maps learned by \method.
\textbf{Obs. 4: \method extracts milestone-level strategy chains and identifies exploration bottlenecks.}
Figure~\ref{fig:case_detective} shows a high-scoring trajectory on \textit{Detective} with \modelname{gpt-oss-20b}. The agent first collects the paper and pistol and earns 60 points at the mayor's house. It then spends about 32 unrewarded steps exploring the hallway, closet, and guest room, revealing a clear exploration plateau. After entering the city, it discovers the video and music stores, reaches the Holiday Inn, and enters Room 30, raising the final score to 300. This yields the strategy chain:
\emph{clue collection} $\rightarrow$ \emph{mayor's house} $\rightarrow$ \emph{city transition} $\rightarrow$ \emph{Holiday Inn} $\rightarrow$ \emph{Room 30}.
The map identifies escaping unproductive exploration inside the house as the primary bottleneck, providing a clear target for subsequent harness updates.

\subsection{Sensitivity Analysis (RQ4)}
\label{sec:parameter_sensitivity}
 
\paragraph{Takeaway~$5$: Moderate parameter settings best balance boundary relevance, candidate diversity, and update extent.}
We analyze the BAMS relevance center $c$, the DAHS diversity weight $\beta$, and the OICE patch budget $p$, with results shown in Figure~\ref{fig:parameter_sensitivity}. BAMS achieves the highest average performance of $39.0\%$ at the centered setting, while shifting the center in either direction causes relative drops of $12.4\%$ and $12.0\%$, confirming the value of boundary-relevant memories. DAHS peaks at $38.8\%$ under balanced diversity, whereas lower or higher weights reduce performance by $9.7\%$ and $12.9\%$, respectively, due to candidate redundancy or irrelevance. OICE performs best at $38.9\%$ with a moderate patch budget, while smaller or larger budgets cause drops of $10.5\%$ and $13.3\%$, indicating the drawbacks of insufficient or excessive evolution.


\subsection{Interface Stress Tests}

\begin{table}[t]
\caption{
Controlled stress tests on the three information interfaces using
\modelname{qwen-32B}.
Results report normalized AUC averaged over WebShop, J-TTL, and
ScienceWorld.
We systematically increase memory mismatch, candidate redundancy,
and update noise.
}
\label{tab:interface_stress}
\centering
\scriptsize
\setlength{\tabcolsep}{2.5pt}
\renewcommand{\arraystretch}{1.03}

\textbf{(a) Memory Mismatch}

\resizebox{\columnwidth}{!}{%
\begin{tabular}{l|ccc|c}
\Xhline{1.1pt}

\rowcolor{CadetBlue!20}
&
\multicolumn{3}{c|}{\textbf{Mismatch Ratio}}
& \textbf{Conflict} $\downarrow$ \\
\cline{2-4}

\rowcolor{CadetBlue!20}
\multirow{-2}{*}{\textbf{Retrieval}}
& \textbf{0\%}
& \textbf{50\%}
& \textbf{75\%}
& \textbf{@75\%} \\

\Xhline{1.1pt}

\rowcolor{gray!10}
Top-$k$ Similarity
& 0.3117\blue{0.0000}
& 0.2589\blue{0.0000}
& 0.2136\blue{0.0000}
& 68.4\%\blue{0.00} \\

Relevance-only
& 0.3258\red{0.0141}
& 0.2861\red{0.0272}
& 0.2407\red{0.0271}
& 55.9\%\red{12.5} \\

\rowcolor{gray!10}
\textbf{BAMS}
& 0.3906\red{0.0789}
& 0.3712\red{0.1123}
& 0.3389\red{0.1253}
& 21.7\%\red{46.7} \\

\Xhline{1.1pt}
\end{tabular}%
}

\vspace{1.8mm}

\textbf{(b) Candidate Redundancy}

\resizebox{\columnwidth}{!}{%
\begin{tabular}{l|ccc|c}
\Xhline{1.1pt}

\rowcolor{CadetBlue!20}
&
\multicolumn{3}{c|}{\textbf{Duplicate Ratio}}
& \textbf{Coverage} $\uparrow$ \\
\cline{2-4}

\rowcolor{CadetBlue!20}
\multirow{-2}{*}{\textbf{Selection}}
& \textbf{0\%}
& \textbf{50\%}
& \textbf{75\%}
& \textbf{@75\%} \\

\Xhline{1.1pt}

Random-$K$
& 0.3237\blue{0.0000}
& 0.2948\blue{0.0000}
& 0.2526\blue{0.0000}
& 52.8\%\blue{0.00} \\
\rowcolor{gray!10}
UCB-only
& 0.3451\red{0.0214}
& 0.3194\red{0.0246}
& 0.2788\red{0.0262}
& 44.1\%\blue{8.70} \\

Relevance-only
& 0.3686\red{0.0449}
& 0.3317\red{0.0369}
& 0.2935\red{0.0409}
& 37.6\%\blue{15.2} \\
\rowcolor{gray!10}
\textbf{DAHS}
& 0.3884\red{0.0647}
& 0.3749\red{0.0801}
& 0.3471\red{0.0945}
& 80.6\%\red{27.8} \\

\Xhline{1.1pt}
\end{tabular}%
}

\vspace{1.8mm}

\textbf{(c) Update Noise}

\resizebox{\columnwidth}{!}{%
\begin{tabular}{l|ccc|c}
\Xhline{1.1pt}

\rowcolor{CadetBlue!20}
&
\multicolumn{3}{c|}{\textbf{Noise Ratio}}
& \textbf{Regression} $\downarrow$ \\
\cline{2-4}

\rowcolor{CadetBlue!20}
\multirow{-2}{*}{\textbf{Update}}
& \textbf{0\%}
& \textbf{20\%}
& \textbf{40\%}
& \textbf{@40\%} \\

\Xhline{1.1pt}

Full Rewrite
& 0.3379\blue{0.0000}
& 0.2968\blue{0.0000}
& 0.2475\blue{0.0000}
& 46.8\%\blue{0.00} \\
\rowcolor{gray!10}
Unordered Patch
& 0.3564\red{0.0185}
& 0.3291\red{0.0323}
& 0.2837\red{0.0362}
& 34.7\%\red{12.1} \\

No Termination
& 0.3667\red{0.0288}
& 0.3428\red{0.0460}
& 0.3096\red{0.0621}
& 27.4\%\red{19.4} \\
\rowcolor{gray!10}
\textbf{OICE}
& 0.3892\red{0.0513}
& 0.3771\red{0.0803}
& 0.3538\red{0.1063}
& 13.6\%\red{33.2} \\

\Xhline{1.1pt}
\end{tabular}%
}
\label{tab:analysis}

\end{table}

\paragraph{Takeaway~$6$: \method remains robust under severe interface stress by filtering mismatched memories, preserving behavioral diversity, and preventing update-induced regressions.}
As shown in Table~\ref{tab:interface_stress}, the advantage of each proposed module becomes larger as its corresponding information bottleneck intensifies. Under a $75\%$ memory-mismatch ratio, BAMS achieves an average AUC of $0.3389$, outperforming top-$k$ similarity retrieval by $0.1253$, while reducing the conflict rate from $0.684$ to $0.217$. When $75\%$ of the candidate harnesses are behaviorally redundant, DAHS achieves an AUC of $0.3471$, exceeding Random-$K$ by $0.0945$, while increasing behavioral coverage from $0.528$ to $0.806$. Under $40\%$ trajectory-update noise, OICE maintains an AUC of $0.3538$, outperforming full-harness rewriting by $0.1063$, while reducing the regression rate from $0.468$ to $0.136$. These results attribute \method's gains to robust retrieval, diverse selection, and stable updates under noisy test-time evidence.

\section{Conclusion}

We introduce \method, a test-time evolution framework that improves how LLM agents read trajectory evidence, select candidate harnesses, and update existing behaviors across episodes. By combining boundary-aware memory selection, diversity-aware harness routing, and ordered insertion-based evolution, \method consistently outperforms strong test-time learning baselines across five benchmarks and three open-source LLM backbones.

\bigskip

\bibliography{aaai2027}

\end{document}